\documentclass[10pt,twocolumn,letterpaper]{article}

\usepackage[utf8]{inputenc} 

\usepackage{iccv}
\usepackage{times}
\usepackage{epsfig}
\usepackage{graphicx}
\usepackage{amsmath}
\usepackage{amssymb}
\usepackage{booktabs}
\usepackage{caption}
\usepackage{soul}
\usepackage{tabularx}
\usepackage[T1]{fontenc}

\usepackage{stackengine}

\usepackage[pagebackref=true,breaklinks=true,letterpaper=true,colorlinks,bookmarks=false]{hyperref}

 \iccvfinalcopy %

\ificcvfinal\fi

\usepackage[table]{xcolor}
\newcommand{\Best}[1]{\textbf{\textcolor{black}{#1}}}
\newcommand{\makecell}[2][@{}c@{}]{\begin{tabular}{#1}#2\end{tabular}}

\DeclareMathOperator*{\argmin}{arg\,min}

\usepackage{overpic}

\newcommand{\overimg}[3][]{%
  \frame{%
    \begin{overpic}[#1]{#2}%
      \put (0, 2) {%
        \setlength{\fboxsep}{2pt}%
        \colorbox{cyan!0!white}{%
          \scriptsize\sffamily\vphantom{y}%
          \vspace{0.2em}%
          #3%
        }%
      }%
    \end{overpic}%
  }%
}

\begin{document}

\title{
Joint Demosaicking and Denoising by Fine-Tuning of Bursts of Raw Images}%

\author{
\begin{tabularx}{\linewidth}{X}
\hfill \makecell{Thibaud Ehret} \hfill \makecell{Axel Davy} \hfill
\hfill \makecell{Pablo Arias} \hfill \makecell{Gabriele Facciolo} \hfill\null\\
\end{tabularx} \\
CMLA, CNRS, ENS Paris-Saclay, Universit\'e Paris-Saclay\\
Universit\'e Paris-Saclay, 94235 Cachan, France\\
{\tt\small thibaud.ehret@ens-cachan.fr}
}

\twocolumn[{%
\renewcommand\twocolumn[1][]{#1}%
\renewcommand\fbox{\fcolorbox{white}{white}}
\maketitle
\begin{center}
{%
\setlength{\fboxsep}{0pt}%
\setlength{\fboxrule}{1pt}%
\stackinset{r}{}{t}{}
{\fbox{\includegraphics[trim={370 260 90 440}, clip, width=0.15\linewidth]{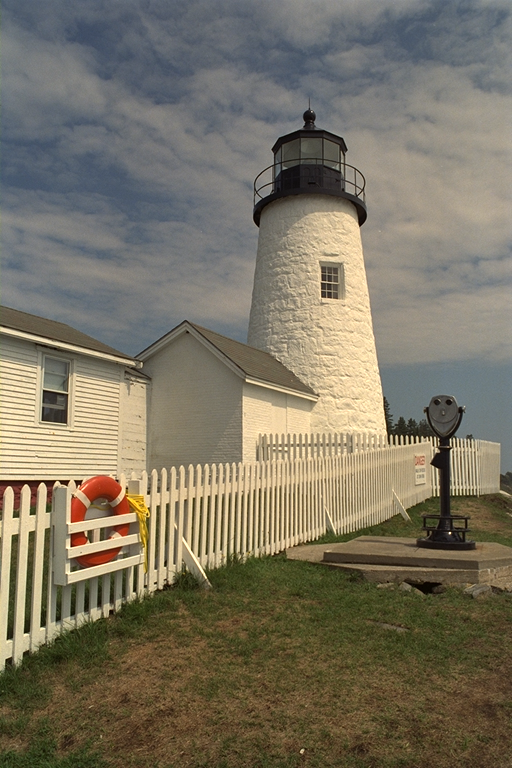}}}
{\overimg[trim={0 30 0 0}, clip, width=0.24\textwidth]{teaser/kodim19.png}{Original}}
\stackinset{r}{}{t}{}
{\fbox{\includegraphics[trim={370 260 90 440}, clip, width=0.15\linewidth]{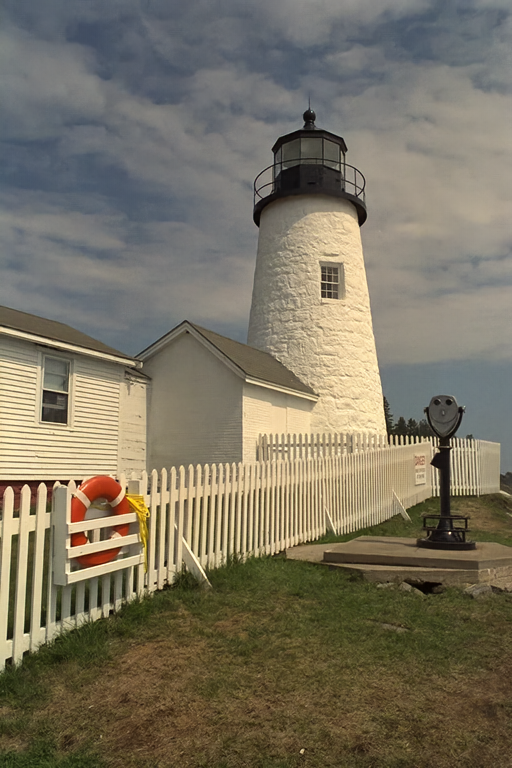}}}
{\overimg[trim={0 30 0 0}, clip, width=0.24\textwidth]{teaser/gharbi.png}{Gharbi \etal: 36.1dB}}
\stackinset{r}{}{t}{}
{\fbox{\includegraphics[trim={370 260 90 440}, clip, width=0.15\linewidth]{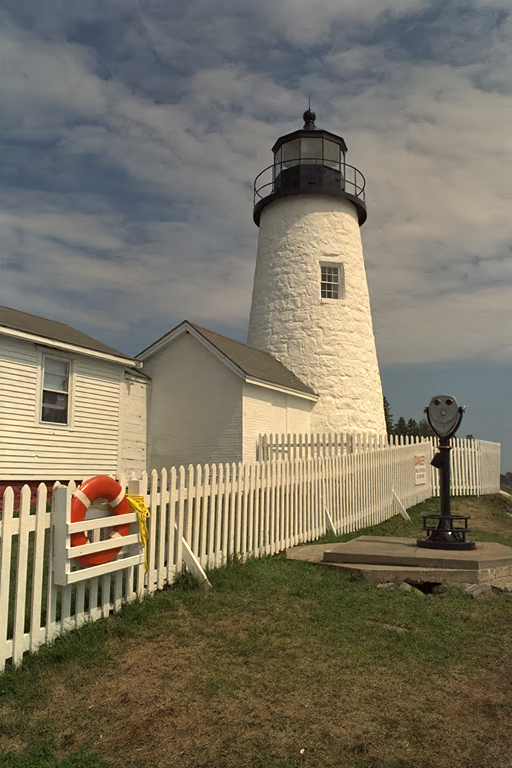}}}
{\overimg[trim={0 30 0 0}, clip, width=0.24\textwidth]{teaser/kokkinos.png}{Kokkinos \etal: 35.5dB}}
\stackinset{r}{}{t}{}
{\fbox{\includegraphics[trim={370 260 90 440}, clip, width=0.15\linewidth]{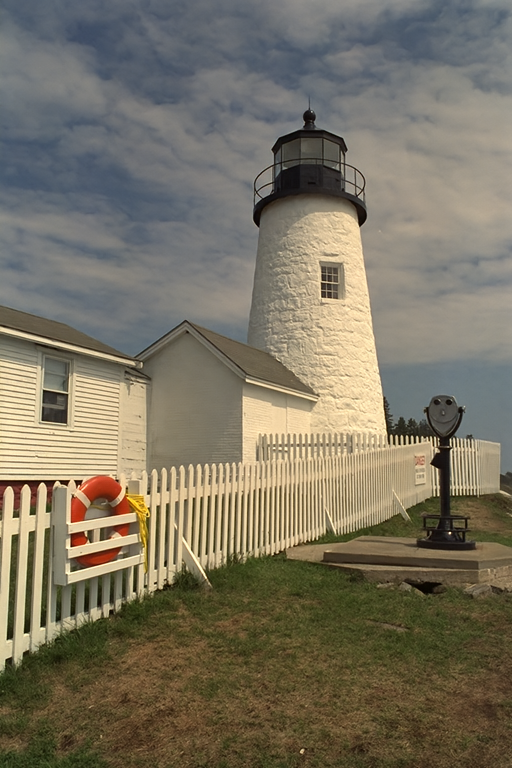}}}
{\overimg[trim={0 30 0 0}, clip, width=0.24\textwidth]{teaser/our_gharbi.png}{Ours: 37.3dB}}

\captionof{figure}{Using a burst, our fine-tuning (starting from the network from Gharbi \etal \cite{gharbi2016deep}) is able to not only denoise well ($\sigma=5$) but also doesn't show any artifacts like {\em zipper} or {\em moire} in the difficult regions. Best visualized on a screen.}
\label{fig:teaser}
}%
\end{center}
}]

\maketitle

\begin{abstract}
   Demosaicking and denoising are the first steps of any camera image processing pipeline and are key for obtaining high quality RGB images. A promising current research trend aims at solving these two problems jointly using convolutional neural networks. Due to the unavailability of ground truth data these networks cannot be currently trained using real RAW images. Instead, they resort to simulated data. 
   In this paper we present a method to learn demosaicking directly from mosaicked images, without requiring ground truth RGB data. We apply this to learn joint demosaicking and denoising only from RAW images, thus enabling the use of real data. In addition we show that for this application fine-tuning a network to a specific burst improves the quality of restoration for both demosaicking and denoising.%
\end{abstract}

\section{Introduction}

Most camera sensors capture a single color at each photoreceptor, determined by a color filter array (CFA) located on top of the sensor. The most commonly used CFA is the so-called Bayer pattern, consisting of a regular subsampling of each color channel. This means, not only that each pixel of the resulting raw image contains one third of the necessary information, but also that the color channels are never sampled at the same positions.
The problem of interpolating the missing colors is called demosaicking and is a challenging ill-posed inverse problem.
To further complicate things, the captured data is contaminated with noise. 

For these reasons the first two steps of a camera processing pipeline are demosaicking and denoising. Traditionally, these problems have been treated separately, but this is suboptimal. Demosaicking first a noisy RAW image 
correlates the noise making its subsequent denoising harder~\cite{park2009denoisingBefore}. Alternatively, if denoising is applied on
the mosaicked data it becomes harder to exploit the cross-color correlations, 
which are useful for color image denoising \cite{dabov2007bm3dcolor,danielyan2009bm3ddemodeno}.

Until recently, state-of-the-art methods for joint demosaicking and denoising  were based on carefully crafted heuristics, such as avoiding interpolation across image edges 
\cite{hirakawa2006demodeno,park2009denoisingBefore,akiyama20154channelDenoising}.
Other methods resort to variational principles where the heuristics are encoded as a prior model~\cite{condat2012tvdemodeno,Heide2014FlexISP}. %
In~\cite{Wronski2019} both problems are addressed simultaneously by aligning and fusing RAW bursts of frames.

Recent data-driven approaches have significantly outperformed traditional model-based methods \cite{khashabi2014randomfields,gharbi2016deep,klatzer2016demodeno,kokkinos2018cascade,kokkinos2018resnet,syu2018learning}. %
In~\cite{gharbi2016deep}, state-of-the-art results are reported with a network trained on a special dataset tailored to demosaicking in which hard cases are over-represented. %
In \cite{kokkinos2018cascade} an iterative neural network is proposed, later improved by \cite{kokkinos2018resnet} obtaining state-of-the-art performance on both real and synthetic datasets.
These networks are relatively lightweight and do not need a lot of training data.  The authors in \cite{syu2018learning} propose two networks for demosaicking. They train on several CFA patterns to compare performance and integrate the handling of denoising with a fine-tuning step.
In~\cite{Zhao2017LossNetworks} the authors find that the artifacts of challenging cases are better dealt with $L_1$ norm, or their proposed combination of the $L_1$ norm with MS-SSIM. 
Meanwhile in~\cite{kwan2019demosackingnasa} alternative metrics to PSNR are also considered. %

The major difficulty in training 
data-driven demosaicking and denoising methods is the difficulty to obtain realistic datasets of pairs of noisy RAW and ground truth RGB images.
For this reason demosaicking networks are trained with simulated data generated by mosaicking  existing RGB images. %
However simulated data follows a statistic that can be different from real data. The RGB images used for training have already been processed by a full ISP (Image Signal Processors) pipeline which includes demosaicking and denoising steps
which leave their footprint on the output image. 
Additionally, the Poisson noise model is only an approximation to the real noise of a specific camera. Several factors can cause deviations. For example the noise can have spatial variations due to temperature gradients in the sensor, or caused by the vignetting or the electronic components in its surroundings.

The need for a specific treatment of realistic noise has been identified in the denoising literature. 
Indeed most of the existing works target synthetic types of noise, e.g. Gaussian noise. Since the noise distribution is well defined, specific methods %
can be crafted \cite{Dabov2006,Lebrun2013,gu2014weighted} and data can be simulated with ground truth so to train neural networks \cite{Zhang2017BeyondDenoising,17-zhang-ffdnet}. However, it has been shown recently in \cite{plotz2017benchmarking} and \cite{SIDD_2018_CVPR} that networks trained on synthetic noise often fail to generalize to realistic types of noise. This has started a trend of study of "real noisy images". For example 
\cite{18-chen-see-in-the-dark,guo2018toward}  
acquire datasets where a low-noise reference image is created by using a longer exposure time. 
Creating this type of dataset is time consuming and prone to bias, as to avoid motion blur in the long exposure the images need to be acquired with a tripod and the scene has to be static.

More recently Lehtinen \etal~\cite{lehtinen2018noise2noise} proposed a novel way of training a denoising network without ground truth, only from pairs of noisy images with independent noise realizations. 
This approach has been taken further by~\cite{noise2void,Batson2019}  which eliminated the need for the second noisy observation, albeit with a penalty in the quality of the obtained results. 
In the context of burst and video denoising the frame-to-frame approach of~\cite{ehret2018model} proposes to fine-tune a pre-trained Gaussian denoising network to other types of noise requiring only a single video. %

\paragraph{Contribution}
In this paper we introduce a mosaic-to-mosaic training strategy analog to the noise-to-noise~\cite{lehtinen2018noise2noise} and frame-to-frame~\cite{ehret2018model} frameworks to be able to handle mosaicked RAW data. 
{The trained network learns to interpolate two thirds of the image data, without having ever seen a complete image.}
This allows us to train both demosaicking and joint demosaicking and denoising networks without requiring ground truth. The resulting networks attain state-of-the-art results, thus eliminating the need to simulate simplistic noise models or to capture time-consuming datasets with long exposure reference frames.
{Although we show results only with a Bayer pattern, our method can equally be applied to other CFA patterns, such as the Fujifilm X-Trans.}
{To the best of our knowledge, this is the first method that learns joint demosaicking and denoising without any ground truth whatsoever; the network has only seen noisy mosaicked images.}

With the proposed framework, we can fine-tune a pre-trained network to a RAW burst. This allows leveraging the already available multi-frame burst data that is present on many mobile camera phones~\cite{Wronski2019}.
The fine-tuning not only adapts the network to the specificities of the camera noise, but it also overfits to the burst. We demonstrate that this overfitting, when controlled, can be beneficial. %
A similar conclusion in the context of single-image super-resolution was reached by the authors of \cite{Shocher2018zero-shot-SR}.
Additionally, when used with an $L_1$ loss, the fine-tuned network naturally handles noise clipping,
a common but challenging problem~\cite{lehtinen2018noise2noise,zhussip2019n2nclipping}.

The proposed strategy can be used to fine-tune other 
demosaicking networks, for example in this paper we show this for the network of \cite{gharbi2016deep} (see Figure~\ref{fig:teaser}), but it could be used in conjunction with more recent burst denoising networks such as \cite{godard2018deep} and \cite{mildenhall2018burst} adapted for CFA images.

The rest of the paper is organized as follows. In Section~\ref{sec:demosaicking_without_gt} we present the proposed mosaic-to-mosaic training of a demosaicking network from a dataset of RAW mosaicked data without ground truth. In Section~\ref{sec:joint_demno} we address the problem of joint demosaicking and denoising given a burst of RAW mosaicked noisy images.
Results are shown in Section~\ref{sec:results}.

\section{Learning demosaicking w/o ground truth}
\label{sec:demosaicking_without_gt}

In this section, we propose a learning method to train demosaicking networks without any ground truth RGB images. Consider two different mosaicked pictures of a same scene $I_1$ and $I_2$. We shall use one image as partial ground truth to learn demosaicking the other (provided that there is a slight movement between the two, so that with high probability the mosaic patterns do not match).

\begin{figure*}
    \centering
    \includegraphics[width=.95\linewidth]{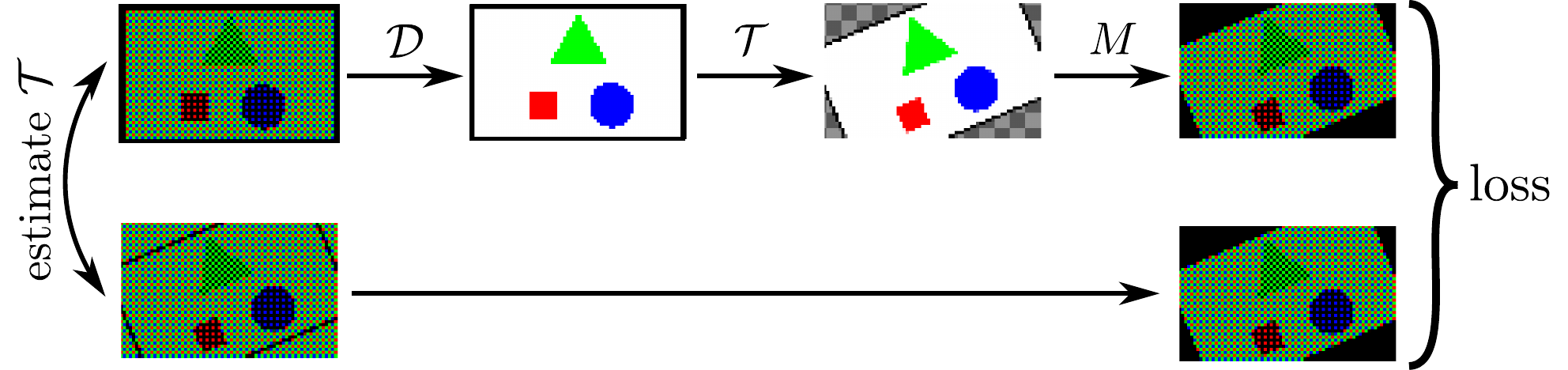}
    \caption{Proposed pipeline to train for demosaicking without using any ground truth. The output after applying the network $\mathcal{D}$ on the first image is warped using the transform $\mathcal{T}$ and masked with $M$ so to be compared to the second masked mosaicked image. The black corners seen in at the last stage of the diagram indicate the undefined pixels after the transform, which are not considered by the loss.}
    \label{fig:b2b_classic}
\end{figure*}

Our method requires that the two pictures can be registered, which is possible when the viewpoints are not too different. This condition is typically met for bursts of images.
Modern cameras systematically take bursts of images, these sequences allow to eliminate shutter lag, to apply temporal noise reduction, and to increase the dynamic range of the device. Nevertheless, the pair of pictures can also be acquired manually by taking two separate pictures of the same scene.

In the following, we suppose we have a set of pairs of images (for example extracted from bursts), where each pair ($I_1$, $I_2$) consists of pictures of the same scene for which we have estimated a transformation $\mathcal{T}$ that registers $I_1$ to $I_2$. In the case of bursts, estimating an affinity is often sufficient. Pairs with not enough matching parts can be discarded. %
The original mosaicked image can be obtained from its demosaicked one by masking pixels.
Thus, if we apply a demosaicking network $\mathcal{D}$ to $I_1$, then apply the transformation $\mathcal{T}$ followed by the mosaicking mask, we are supposed to get $I_2$.
We can compute $M(\mathcal{T}(\mathcal{D}(I_1)))$, where $M$ represents the mosaicking operation (masking pixels), compute a distance to $I_2$, which acts as ground truth,
and backpropagate the gradient to train $\mathcal{D}$. In some sense, $I_2$ acts as a partial ground truth, as only one third of $\mathcal{T}(\mathcal{D}(I_1))$ gets compared to $I_2$. However, contrary to artifical RGB ground truths, we do not suffer from bias introduced by the RGB processing pipeline, nor require complex settings to produce these RGB ground truths. We implemented $\mathcal{T}$ with a bicubic interpolation through which gradient can be backpropagated.  %
This results in the following loss:

\begin{equation}\label{eq:main_equation}
\ell_p(\mathcal D(I_1),I_2) = \|M(\mathcal{T}(\mathcal{D}(I_1))) - I_2\|_p^p,
\end{equation}
where $p = 1,2$. The norm is computed only in the pixels where both images are defined.
In this section we use $p=2$ (squared $L_2$ norm).
The training method is depicted in Figure~\ref{fig:b2b_classic}.  Our learning process can  be linked with~\cite{lehtinen2018noise2noise,Batson2019}. The main difference is that we have an \textit{a priori} on the position of the degraded pixels.

\begin{figure*}
    \centering
    \includegraphics[width=1\linewidth]{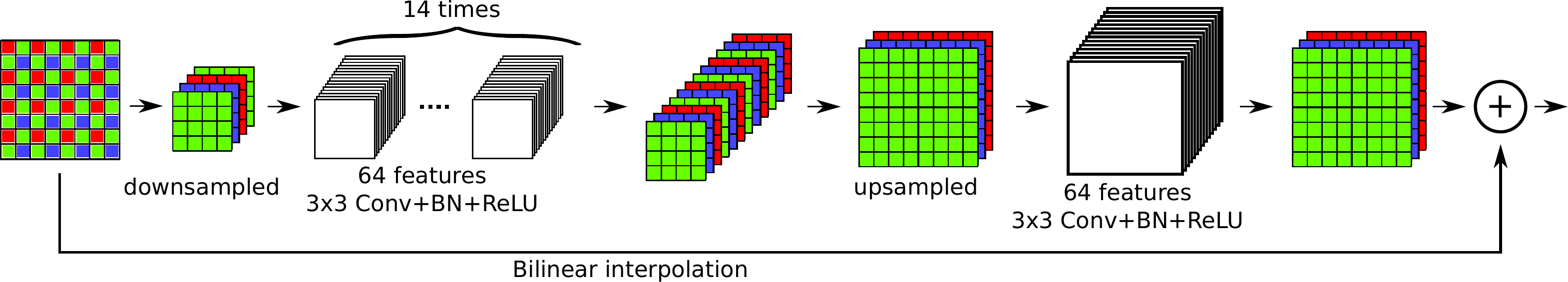}
    \caption{Architecture of the network used to compare the performance of learning on RGB ground truth or only with pairs of RAW images.}
    \label{fig:arch_dmcnn}
\end{figure*}

\paragraph{Demosaicking network} To test the proposed training, we will use in this section a network architecture heavily inspired by the one from Gharbi \etal~\cite{gharbi2016deep} while using improvements suggested in more recent work with the usage batch normalization layers~\cite{Ioffe2015BatchShift} as well as residual learning~\cite{He2016}. These techniques are known to speed-up training time and sometimes increase performance. %
The network starts with a four-channel Bayer image that goes through a series of $14$ Conv+BN+ReLu layers with $64$ features and $3\times 3$ convolutions. A $15th$ layer of Conv+BN+ReLu produces $12$ features with $3\times3$ convolutions. It is followed by an upsampling layer producing an RGB image of twice the width and twice the height. Like Gharbi et al. we added a layer (a Conv+BN+ReLu with $3\times3$ convolutions) before the layer producing the final output. Since our network is residual we need to add the bilinearly interpolated RGB image to produce the final result. All convolution layers have padding to keep the resolution constant from beginning to end. %
The architecture of the network is depicted in Figure~\ref{fig:arch_dmcnn}. %

\paragraph{Comparing learning with ground truth RGB and our method}
We verify that this method for training demosaicking without ground truth is competitive with classic supervised training by training the \emph{same} architecture with both methods and show comparable results. For this experiment we considered a mosaicking with Bayer pattern which is the most frequent mosaicking pattern. 

In order to be able to compare the results of training with and without ground truth, we decided to simulate the pairs on which the demosaicking is trained. For both trainings we use the dataset of \cite{syu2018learning}, which consists of 500 images (of sizes around $700\times500$) from Flickr. To generate pairs to learn with our method, we warped the same RGB image with a random affinity - thus simulating two views - and generated the mosaicked images from them. To speed-up the training we chose the same transform for all patches of a same batch. We trained both networks for $45$ epochs using Adam and a learning rate of $10^{-2}$. We reduced the learning rate by a factor of $10$ at epochs $20$ and $40$.

\begin{figure}
    \centering
    \includegraphics[width=.9\linewidth]{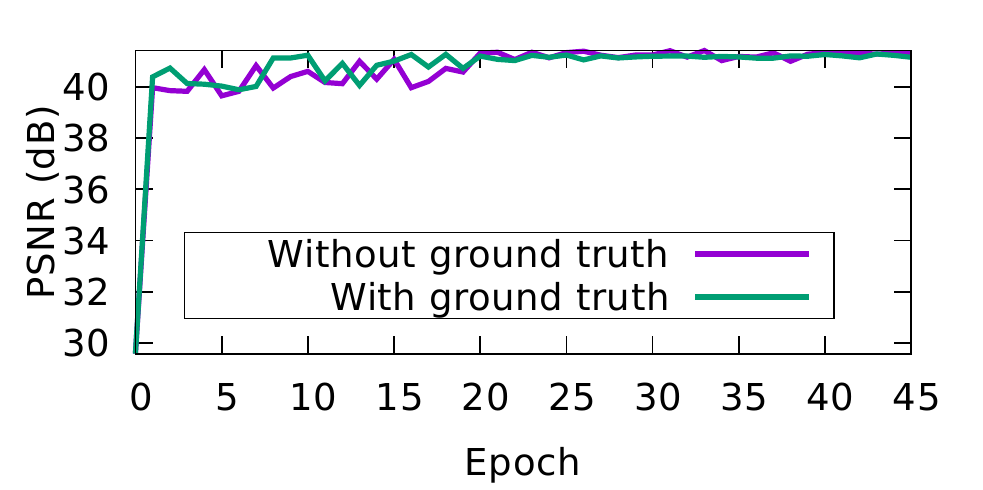}
    \caption{Evolution of the average PSNR on the Kodak dataset when training with ground truth data and when training without RGB ground truth data available. Training without RGB ground truth behaves the same than training with an RGB ground truth.}
    \label{fig:learning_curve_b2b_classic}
\end{figure}

Figure \ref{fig:learning_curve_b2b_classic} compares the evolution of the PSNR on the Kodak dataset\footnote{http://r0k.us/graphics/kodak/} while training our network with ground truth against the training without ground truth. It can be observed that training without ground truth behaves the same as with the ground truth. The convergence speed seems to be equivalent as well as the final demosaicking quality.

Table~\ref{tab:b2b_classic} shows the quality of demosaicking using either ground truth or no ground truth versus the state of the art in image demosaicking. The model learned without having ever seen an RGB image is able to achieve the same quality than the same network trained using the RGB ground truth, which indicates that having a ground truth is not necessary to obtain state-of-the-art performance on this task. For comparison, we also show the results obtained with model-based methods \cite{getreuer2011color,Heide2014FlexISP} that do not need training with ground truth (they do not need training at all).
\begin{table}
	\begin{center}
		{\small
		\renewcommand{\tabcolsep}{1.6mm}
        \renewcommand{\arraystretch}{1.3}
        \begin{tabular}{ @{}l c c @{}}
            \toprule
            Method                                     & With ground truth & Without ground truth    \\
            \midrule
            Getreuer \etal~\cite{getreuer2011color}   & -               & 38.1 \\
            Heide \etal~\cite{Heide2014FlexISP}      & -               & 40.0 \\
            Gharbi \etal~\cite{gharbi2016deep}        & 41.2            & -    \\
            Network from~\S\ref{sec:demosaicking_without_gt}	                                   & 41.2            & 41.3 \\
            \bottomrule
        \end{tabular}}
    \end{center}
    \caption{PSNR results for different demosaicking methods on the Kodak dataset. Training without ground truth (network from \S\ref{sec:demosaicking_without_gt}) outperforms the methods without ground truth while still achieving state-of-the-art PSNRs.}
	\label{tab:b2b_classic}
\end{table}

\section{Joint demosaicking and denoising by fine-tuning on a burst}
\label{sec:joint_demno}

In the previous section, we demonstrated that having a training dataset with RGB ground truth is not mandatory to reach state-of-the-art performance: Similar demosaicking performance is reached with a just a database of pairs of RAW mosaicked data. While this was demonstrated on noise-free images, it can also be done when images are noisy. In this section we show an application of the method to online fine-tuning on bursts. We present the method on two networks: the noiseless network from Section \ref{sec:demosaicking_without_gt} where the network has to learn to denoise using only the burst (this is a sort of a toy example) and the state-of-the-art network from \cite{gharbi2016deep}.

\paragraph{Joint demosaicking and denoising without ground truth}

\begin{figure*}
    \centering
    \includegraphics[width=0.33\linewidth]{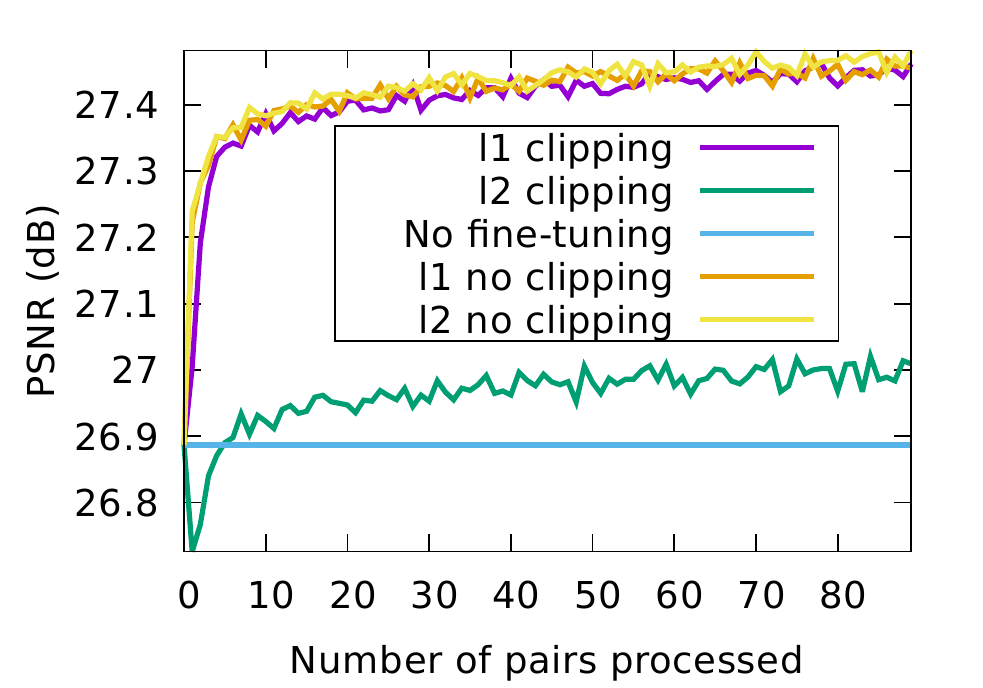}
    \includegraphics[width=0.33\linewidth]{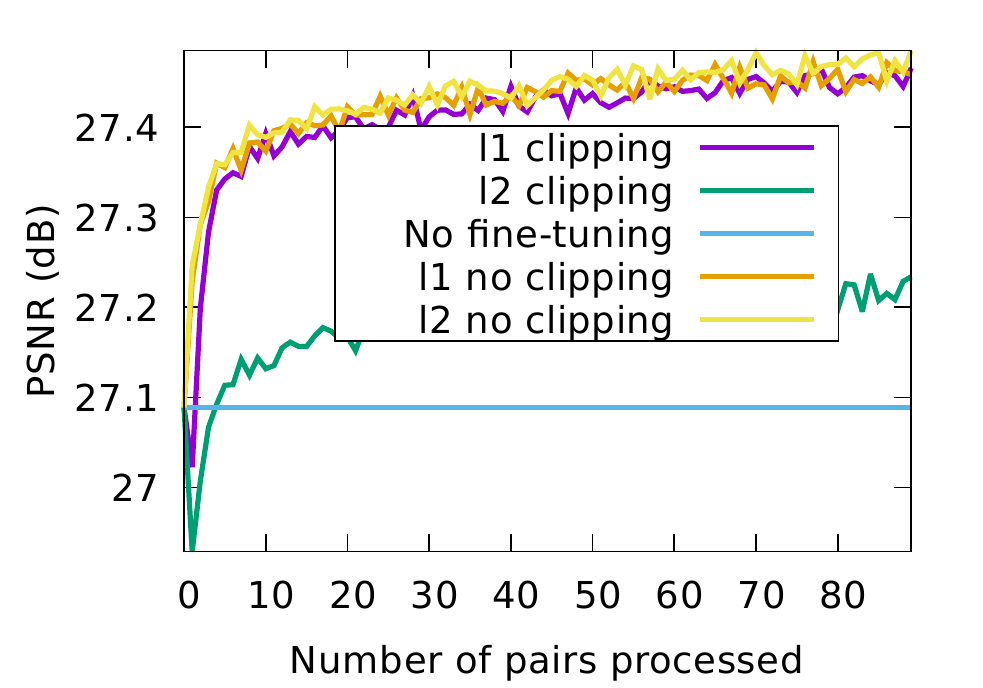}
    \includegraphics[width=0.33\linewidth]{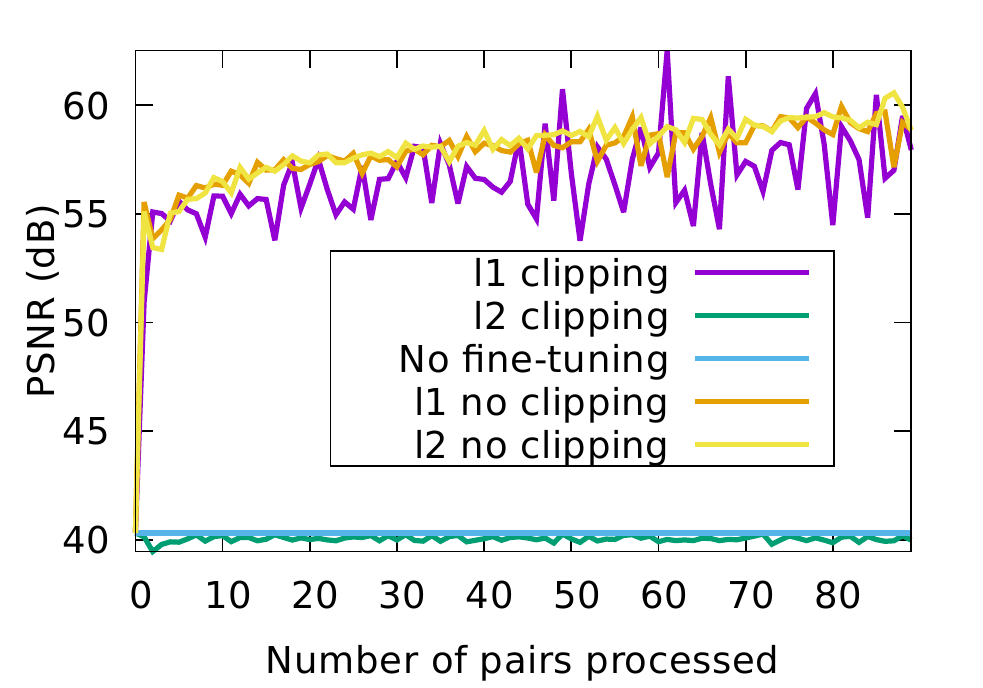}
\caption{Fine-tuning a denoising network (DnCNN~\cite{Zhang2017BeyondDenoising} $\sigma=25$) on a burst of 10 noisy ($\sigma=25$) grayscale images with saturated regions. From left to right: PSNRs over the whole image, on non-saturated regions, and on the saturated regions. After fine-tuning, the network works better both on saturated and non-saturated regions. While $L_2$ is not able to deal with clipping, using $L_1$ for fine-tuning performs similarly to fine-tuning without clipping. %
    }
    \label{fig:expe_clipping}
\end{figure*}

Using the noise-to-noise framework presented in~\cite{lehtinen2018noise2noise}, we aim to train a network with parameters $\theta$. Supervised learning of a joint demosaicking and denoising network corresponds to solving
\begin{equation}
\argmin_\theta \sum_i L(f_\theta(x_i), y_i),
\end{equation}
where the $x_i$ are noisy mosaicked images, and the $y_i$ are their ideal noise-free demosaicked images, $L$ is a loss such as $L_2$ or $L_1$.
In the noise-to-noise framework, the equivalent problem (conditionally on the noise being mean preserving for $L_2$, or median preserving for $L_1$) is to solve
\begin{equation}
\argmin_\theta \sum_i L(f_\theta(x_i), \hat{y}_i),
\end{equation}
where $\hat{y}_i$ are noisy observations of $y_i$.

Combining this with Equation~\eqref{eq:main_equation}, our proposal is to solve
\begin{equation}
\argmin_\theta \sum_i \ell_1(f_\theta(x_i), z_i),
\label{eq:overfitting_l1}
\end{equation}
where the $(x_i, z_i)$ are pairs of noisy images of the same scene, and $\ell_p$ was introduced in Section~\ref{sec:demosaicking_without_gt}. We use $p=1$ in this section ($L_1$ norm), which allows to handle clipped noise (see discussion on the choice of the loss).

The loss requires the computation of a transform $\mathcal T$ matching each pair of mosaicked images. For that we use the inverse compositional algorithm \cite{thevenaz1998pyramid,baker2001equivalence} to estimate a parametric transform (in practice we estimate an affinity which we found to be well-suited for bursts). An implementation of this method is available in~\cite{briand2018}. The advantage of this method is that it is robust to noise and can register two images very precisely (provided that they can be registered with an affinity). Since we only have access to Bayer images of size $W\times H$, the first step is to generate four-channel images of size $\frac{W}{2}\times \frac{H}{2}$ corresponding to the four phases of the Bayer pattern. The transform is then estimated on these images before upscaling it to the correct size. 

Having the pairs with the associated transform, one can finally apply the pipeline presented in Section~\ref{sec:demosaicking_without_gt} and in Figure~\ref{fig:b2b_classic}. As in~\cite{ehret2018model} we initialize the network using a pretrained one. In particular in the following, we use the network trained for demosaicking without ground truth presented in Section~\ref{sec:demosaicking_without_gt}, as well as the network from~\cite{gharbi2016deep}. %

\vspace{-1em}

\paragraph{Choice of Loss}

One particularly well known problem with denoising is clipped noise: The underlying signal $I$ belongs to a fixed range, but the noise can make it leave that intensity range. Due to hardware clipping, the measured image is inside the fixed range, and thus the noise statistics are biased. When minimizing with the $L_1$ norm over the same image with several noise realizations, the best estimator is the median of the realizations~\cite{lehtinen2018noise2noise}, which is unaffected by the hardware clipping. Thus by using $L_1$ norm and fine tuning on a burst, our method handles clipping without any pre or post-processing required. This phenomenon is illustrated in Figure~\ref{fig:expe_clipping} with a classic denoising network, DnCNN~\cite{Zhang2017BeyondDenoising}.

\begin{figure*}
    \centering
    \includegraphics[trim={100 375 550 23}, clip, width=0.24\linewidth]{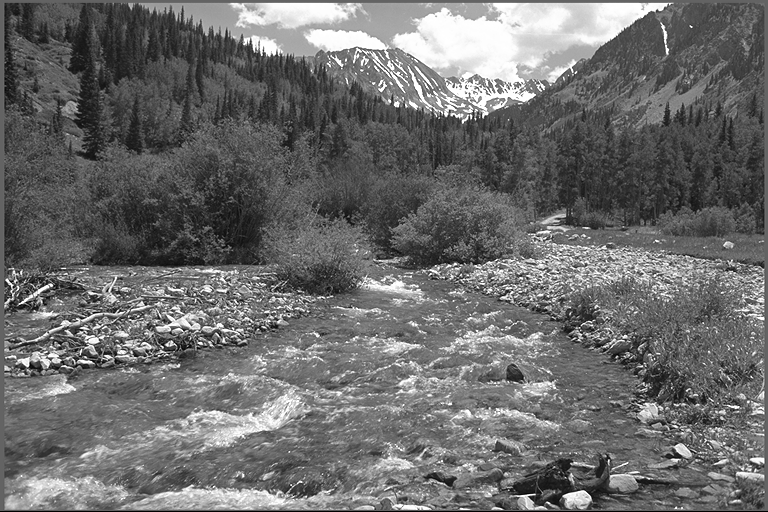}
    \includegraphics[trim={100 375 550 23}, clip, width=0.24\linewidth]{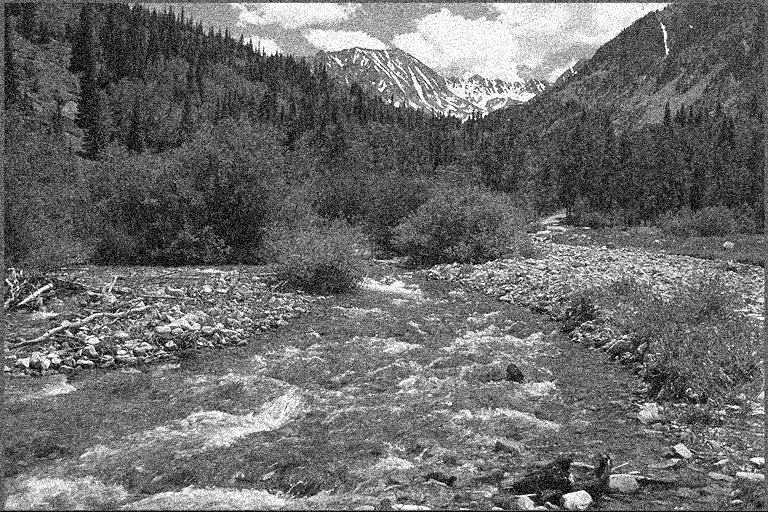}
    \includegraphics[trim={100 375 550 23}, clip, width=0.24\linewidth]{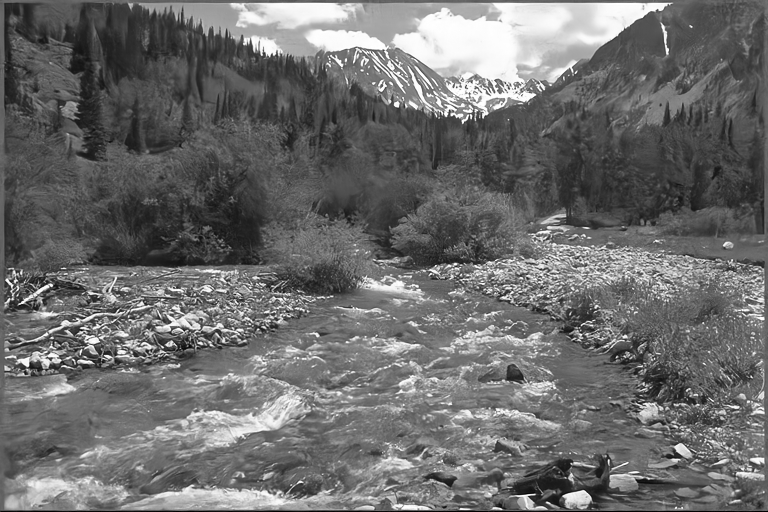}
    \includegraphics[trim={100 375 550 23}, clip, width=0.24\linewidth]{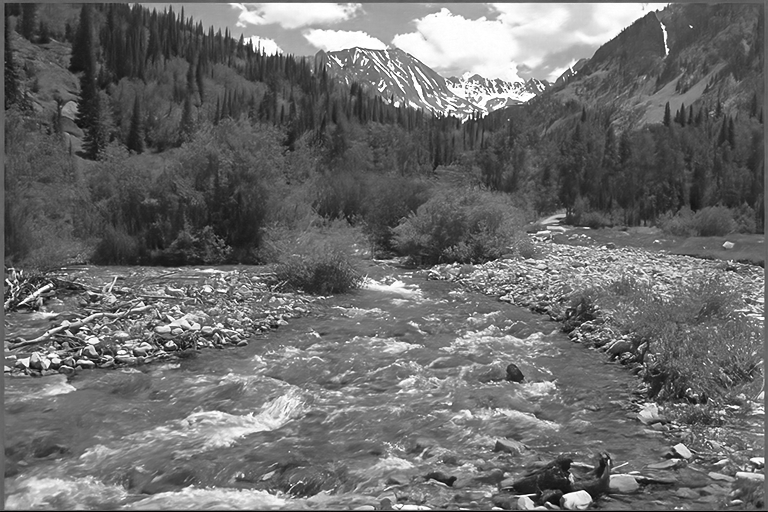}
    \caption{From left to right: reference image, noisy ($\sigma=25$), pretrained DnCNN and DnCNN after fine-tuning. The details, such as the trees, are sharper and more distinguishable after fine-tuning.  Figure best visualized zoomed-in on a computer.}
    \label{fig:expe_overfitting_noise_visual}
\end{figure*}

\begin{figure}
    \centering
    \includegraphics[width=0.39\linewidth]{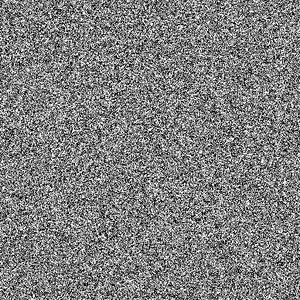}
    \includegraphics[width=0.39\linewidth]{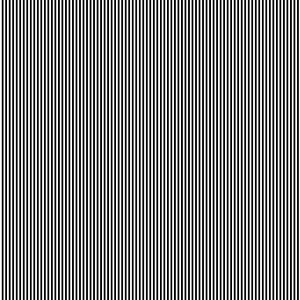}
    \includegraphics[width=.9\linewidth]{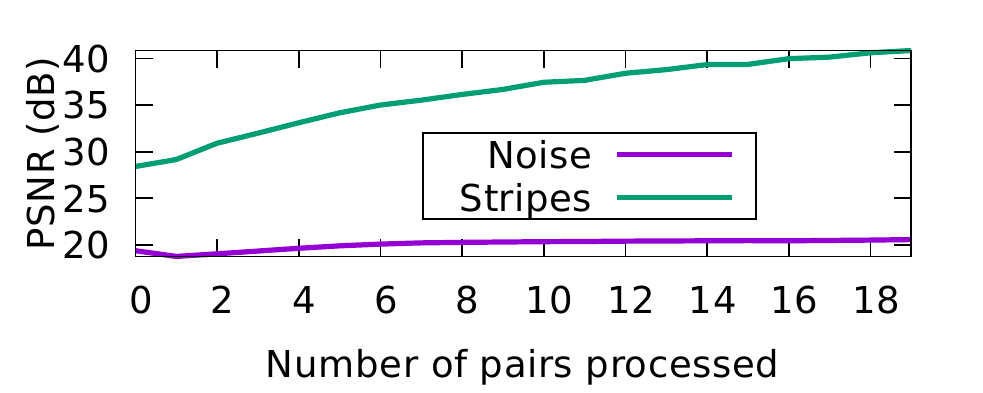}
    \caption{From left to right: image of binary noise and an image of stripes. Fine-tuning DnCNN on the very self-similar image of stripes leads to a much bigger increase in quality compared to the image of binary noise.}
    \label{fig:expe_similarity}
\end{figure}

\vspace{-1em}
\paragraph{Fine-tuning to a single scene}
By fine-tuning over a single burst the network ends up overfitting the data. Usually overfitting to the training data is avoided as it results in a poor generalization ability.
However, in our case the fine-tuned network will only be applied to that burst, and overfitting improves the result for that specific burst.
There are other examples in the literature where a network is overfitted to a specific input (or a small dataset of inputs). For example, \cite{caelles2017osvos} turns an object classification network into a video segmentation one by fine-tuning it on the first frame, which is labeled. The network then learns to track the labeled objects in the following frames. Several image restoration problems are addressed in~\cite{ulyanov2018deep} by using a network as a prior. The network parameters are trained for each input image. In~\cite{Shocher2018zero-shot-SR} a super-resolution CNN is trained by fine-tuning the specific structures of the current image. In~\cite{ehret2018model} a pre-trained denoising is fine-tuned to an input video. These can be related to recent works on meta-learning~\cite{finn2017metalearning}.

This fine-tuning is also reminiscent of traditional image processing methods that fit a model to the patches of the image. In~\cite{yu2012ple} the image patches are modeled using a Gaussian mixture model (GMM), in~\cite{Elad2006} by representing them sparsely over a learned dictionary, and in~\cite{morup2008csc} via sparse convolutions over a set of kernels. In all these cases the models were trained on the input image. The assumption underlying these methods is that images are self-similar and highly redundant, allowing for compact representations of their patches. %

Figure~\ref{fig:expe_overfitting_noise_visual} shows that fine-tuning a grayscale denoising network (DnCNN) on a burst of images can significantly improve the denoising results. The likely explanation is that the network is able to capture a part of the image self-similarity, similar to the model-based methods. Figure \ref{fig:expe_similarity} illustrates the performance evolution when fine-tuning a denoising network on a set of noisy realizations of two synthetic images, one of stripes (thus very self-similar) and a binary noise image (thus not self-similar). The performance gap is explained by the self-similarity of the former image. %

\section{Experimental results}
\label{sec:results}

{To evaluate quantitatively the performance of the proposed training strategy, we first apply it on simulated data, since there are no real noisy raw bursts with ground truth publicly available. We generate the burst from a single image by applying random affinities. In the cases where noise is considered, the added noise is white Gaussian. During training, the affinities are estimated from the noisy raw data.} Code to reproduce the results is available on \url{https://github.com/tehret/mosaic-to-mosaic}

\vspace{-1em}

\paragraph{Network fine-tuning on a sequence outperforms single image denoising}

\begin{figure}
    \centering
    \includegraphics[width=.85\linewidth]{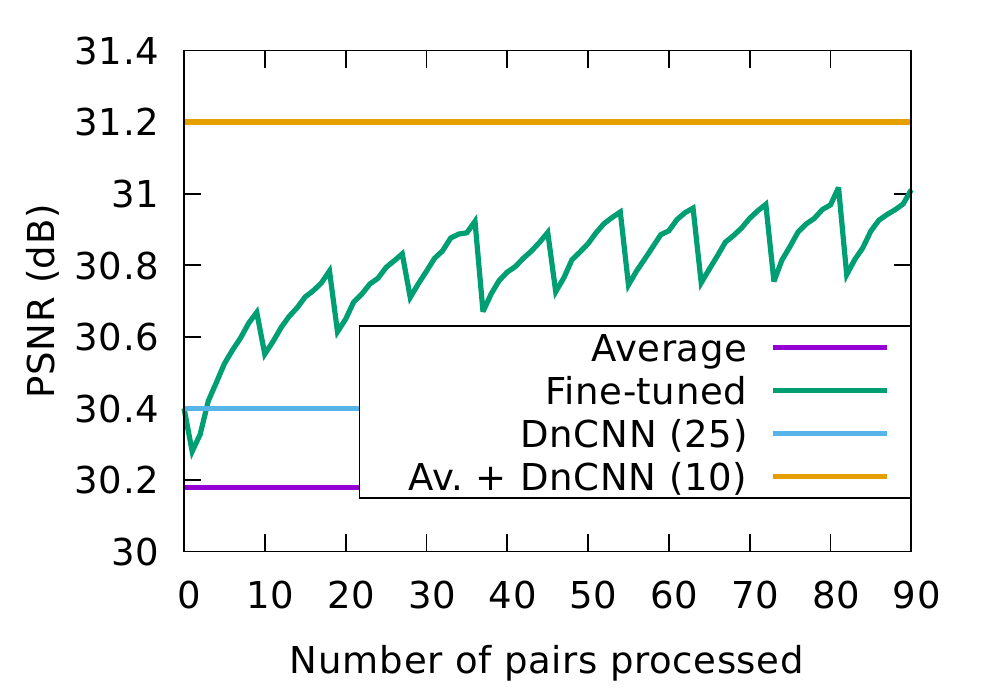}
    \caption{Fine-tuning a pre-trained denoising network (DnCNN $\sigma=25$)  to a specific sequence increase the quality of the result. The visible drops correspond to each change of image considered (pairs are considered in lexicographical order). It is important to finish with the reference image as to maximise the performance. 
    The fine-tuned network, which takes as input a single frame, comes close to the performance of DnCNN applied to the temporal average of all frames.}
    \label{fig:expe_overfitting_noise}
\end{figure}

Fine-tuning a network to a sequence allows to restore the image beyond the performance of a single image denoising.
In the experiment shown in Figure~\ref{fig:expe_overfitting_noise} a sequence of 10 frames without mosaicking pattern nor motion is considered. The plot shows the PSNR evolution as the fine-tuning processes all the pairs (90 in total). 

We consider the pairs in lexicographical order, that is every time a new input image is selected it is sequentially paired with all other images in the sequence.
Note the characteristic shape traced by the PSNR curve: every time a new input image is selected the performance first drops and then steadily improves surpassing the previous peak. This shows that not only the network is adapting the current input image but it is also building upon previously seen images.

This fine tuning can be linked to a temporal noise reduction (TNR). For comparison the plot includes  the PSNR of results obtained by averaging the frames, which amounts to a naive TNR, by denoising a single frame with DnCNN, and by denoising the naive TNR result with DnCNN. The latter amounts to the best possible TNR result in this ideal case. Note that the fine tuning is largely surpassing the performance of  single image denoising and  approaches TNR with DnCNN.
In practice temporal averaging followed by denoising is not as straightforward on mosaicked images, so there is no equivalent of this upper bound on mosaicked images. This justifies the relevance of the proposed method. %

\vspace{-1em}

\begin{figure*}
    \centering
    \includegraphics[width=0.24\linewidth]{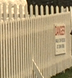}
    \includegraphics[width=0.24\linewidth]{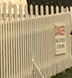}
    \includegraphics[width=0.24\linewidth]{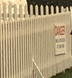}
    \includegraphics[width=0.24\linewidth]{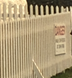}
    \caption{From left to right: reference image, the network from \S\ref{sec:demosaicking_without_gt}, the network from \S\ref{sec:demosaicking_without_gt} after fine-tuning and Gharbi \etal. Because of the reduced size of the training set our blind network still has some {\em moire} artifact but they completely disappear after fine-tuning on the data achieving a result visually close to Gharbi \etal without having to learn on a specific well-chosen dataset. Figure best visualized zoomed-in on a computer.}
    \label{fig:expe_overfitting_demosaicking_visual}
\end{figure*}

\begin{figure}
    \centering
    \includegraphics[width=.85\linewidth]{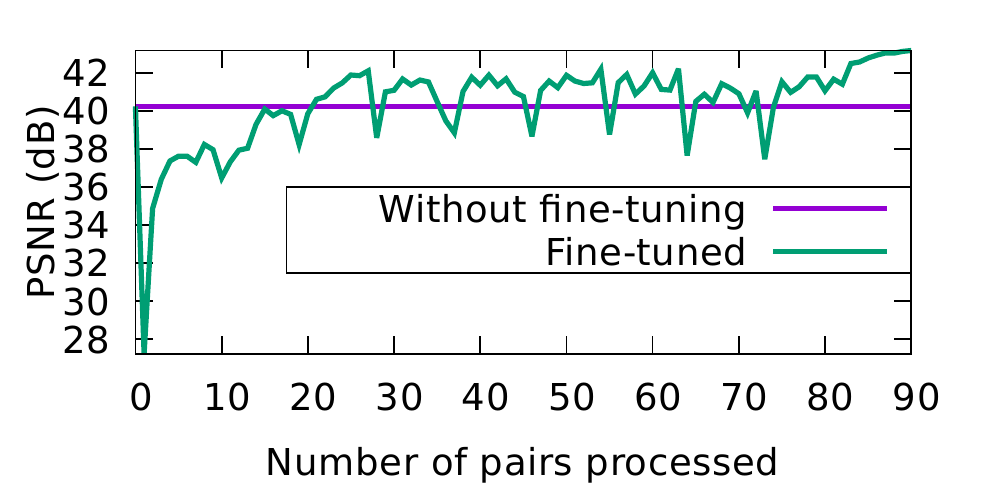}
    \caption{Fine-tuning a pre-trained demosaicking network (from Section \ref{sec:demosaicking_without_gt}) to a specific sequence increase the quality of the result. The visible drops correspond to each change of image considered (pairs are considered in lexicographical order). It is important to finish with the reference image as to maximise the performance.}
    \label{fig:expe_overfitting_mosaic}
\end{figure}

\paragraph{Improving demosaicking by fine-tuning} 
Similarly to denoising, fine-tuning improves demosaicking. The evolution of the improvement, showed in Figure~\ref{fig:expe_overfitting_mosaic}, is quite similar to the one presented for denoising. Moreover artifacts that existed in the initial network, due to a low amount of training, are removed completely by the fine-tuning, see Figure \ref{fig:expe_overfitting_demosaicking_visual}. The result then looks visually very similar to the result from Gharbi \etal that was trained specially to deal with these difficult cases.

Table~\ref{tab:overfitting_demosaicking} compares the PSNR obtained for different networks on the Kodak dataset. The network from Section \ref{sec:demosaicking_without_gt} was fine-tuned on {\em kodim19}, which is singled out in the table. As expected, the fine-tuned network works well on the reference image but its performance decreases on the other images. The network without fine-tuning performs better on the whole Kodak dataset than the network that was fine-tuned on a specific image. The increase in performance for this reference image after fine-tuning was of more than~$2dB$.

\begin{table}
	\begin{center}
		{\small
		\renewcommand{\tabcolsep}{1.6mm}
        \renewcommand{\arraystretch}{1.3}
        \begin{tabular}{ @{}l c c @{}}
            \toprule
            Method      & kodim19 & Kodak dataset    \\
            \midrule
            \S\ref{sec:demosaicking_without_gt} fine-tuned on kodim19  & \Best{44.4}      & 40.4 \\
            \S\ref{sec:demosaicking_without_gt} without fine-tuning    & 42.1             & \Best{41.3} \\
            \bottomrule
        \end{tabular}}
    \end{center}
    \caption{PSNR results using an fine-tuned network on the lighthouse image of the Kodak dataset (kodim19) versus the same network without fine-tuning. While fine-tuning improves on the specific image, the overall performance on the dataset is decreased.}
    \label{tab:overfitting_demosaicking}
\end{table}

\vspace{-1em}
\paragraph{Joint demosaicking and denoising using fine-tuning}
The final application of fine-tuning is to do both previous applications at the same time. Table~\ref{tab:joint} compares the result to two other methods of joint demosaicking and denoising. The networks were fine-tuned on each image individually. Overall this approach is very competitive. A network that had never seen noise before (\S\ref{sec:demosaicking_without_gt} + our fine-tuning) is now able to perform at the same level as one of the best network trained for this specific application. When using the state-of-the-art network from~\cite{gharbi2016deep}, our fine-tuning improves the final quality by more than $1dB$.

Not only do we achieve competitive results in terms of PSNR, the results are free of demosaicking artifacts.
Indeed, as shown in Figure~\ref{fig:teaser}, even in the regions that are particularly hard such as the fence. For example there is no \em zippering compared to the result of Gharbi~\etal. 

\begin{table}
	\begin{center}
		{\small
		\renewcommand{\tabcolsep}{1.6mm}
        \renewcommand{\arraystretch}{1.3}
        \begin{tabular}{ @{}p{1.4cm} c c c c @{}}
            \toprule
            Method     & \cite{gharbi2016deep} & \cite{kokkinos2018resnet} & \shortstack{\S \ref{sec:demosaicking_without_gt} + our\\ fine-tuning} &  \shortstack{\cite{gharbi2016deep} + our\\ fine-tuning}  \\
            \midrule
            01, $\sigma\!=\!5$   & 34.9/.9584  & 34.5/.9540 & 35.1/.9545 & \Best{35.9}/\Best{.9633} \\
            13, $\sigma\!=\!5$   & 32.9/.9574  & 32.3/.9515 & 33.6/.9587 & \Best{34.3}/\Best{.9641} \\
            16, $\sigma\!=\!5$   & 37.1/.9496  & 36.5/.9390 & 36.1/.9399 & \Best{38.2}/\Best{.9570} \\
            19, $\sigma\!=\!5$   & 36.1/.9430  & 35.5/.9380 & 36.3/.9375 & \Best{37.3}/\Best{.9500} \\
            All, $\sigma=5$     & 36.2/.9465  & 35.2/.9329 &  36.0/.9401 & \Best{37.6}/\Best{.9559}  \\
            \midrule
            19, $\sigma\!=\!10$  & 33.2/.8958  & 31.1/.8612 & 32.9/.8877 & \Best{34.0}/\Best{.9067} \\
            19, $\sigma\!=\!10$ 20 images & 33.2/.8958  & 31.1/.8612 & 33.2/.8935 & \Best{34.3}/\Best{.9091} \\
            \bottomrule
        \end{tabular}}
    \end{center}
    \caption{PSNR results of different methods for the task of joint demosaicking and denoising. It shows that even though our method is completely blind, it is able to compete with the state of the art. The rows identify different images from the Kodak dataset, and noise levels. Moreover increasing the length of the burst also allows to improve the quality in the cases where it might perform worse otherwise. Our method used the network trained in Section \ref{sec:demosaicking_without_gt} and was fine-tuned with $10$ generated noisy images except when mentioned otherwise.}
	\label{tab:joint}
\end{table}

\begin{figure}
    \centering
    \includegraphics[width=0.46\linewidth]{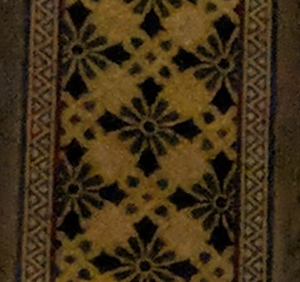}
    \includegraphics[width=0.46\linewidth]{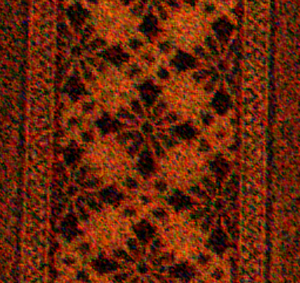}
    \includegraphics[width=0.46\linewidth]{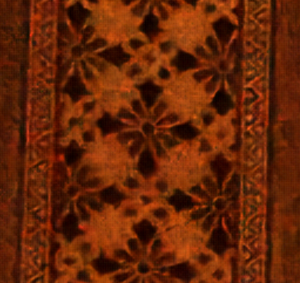}
    \includegraphics[width=0.46\linewidth]{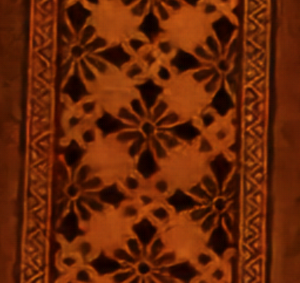}
    \caption{Experiment on a real burst. Top left to bottom right: The result of the HDR+ pipeline~\cite{Hasinoff2016HDRplus}, bilinear interpolation, \cite{gharbi2016deep} and \cite{gharbi2016deep} with our fine-tuning. 
    Contrast was enhanced for all methods except HDR+. Note that the HDR+ pipeline includes color balance as well as sharpening. It also uses all the images of the burst to produce the result (all other methods use only the reference frame). Figure best visualized zoomed in on a computer.}
    \label{fig:real}
\end{figure}

The final experiment is on real data. We took a burst from the HDR+ dataset~\cite{Hasinoff2016HDRplus} and applied our process. We compare the result of a simple bilinear interpolation, the result of \cite{gharbi2016deep} and \cite{gharbi2016deep} with our fine-tuning in \ref{fig:real}. Fine-tuning allows for a better denoising and a better reconstruction of details while limiting artifacts.

\vspace{-1em}
\paragraph{Remarks on computation cost}

We empirically found that the amount of data needed for fine-tuning the network is linked to the number of pixels and not the number of images of a burst. 
This allows to fine-tune even on short bursts like the ones from the HDR+ dataset (of size $2400\times 1300$) using at most six images.
Regarding computation time, we presented fine-tuning as an offline application, for example for professional photography where best quality is required. However, recent works \cite{Tonioni2019Learning,Tonioni2018RealtimeStereo} have shown that fine-tuning can also be achieved in real time for videos.

\section{Conclusion}

In this work, we have proposed a novel way of training demosaicking neural network without any RGB ground truth, by using instead other mosaicked data of the same scene (such as from a burst of images). Based on it and on recent neural network advances, we proposed a method to train jointly demosaicking and denoising with bursts of noisy raw images. We show that fine-tuning on a given burst boosts the reconstruction performance. Clipped noise, a hard problem, is handled natively. It also presents a specific case where overfitting a network to the training data is valuable. Since we do not expect generalization there's only benefits from this overfitting.

We hope our work can lead to new camera pipeline calibration procedures, and general improvement of the image quality when a burst is available.

\section*{Acknowledgements}
The authors gratefully acknowledge the support of NVIDIA Corporation with the donation of the Titan V GPU used for this research. Work partly financed by IDEX Paris-Saclay IDI 2016, ANR-11-IDEX-0003-02, Office of Naval research grant N00014-17-1-2552, DGA Astrid project \guillemotleft filmer la Terre\guillemotright~n$^o$ ANR-17-ASTR-0013-01, MENRT.

{\small
\bibliographystyle{ieee_fullname}
\bibliography{egbib}

\begin{thebibliography}{10}\itemsep=-1pt

\bibitem{SIDD_2018_CVPR}
Abdelrahman Abdelhamed, Stephen Lin, and Michael~S. Brown.
\newblock A high-quality denoising dataset for smartphone cameras.
\newblock In {\em The IEEE Conference on Computer Vision and Pattern
  Recognition (CVPR)}, June 2018.

\bibitem{akiyama20154channelDenoising}
Hiroki Akiyama, Masayuki Tanaka, and Masatoshi Okutomi.
\newblock Pseudo four-channel image denoising for noisy cfa raw data.
\newblock In {\em 2015 IEEE International Conference on Image Processing
  (ICIP)}, pages 4778--4782. IEEE, 2015.

\bibitem{baker2001equivalence}
Simon Baker and Iain Matthews.
\newblock Equivalence and efficiency of image alignment algorithms.
\newblock In {\em IEEE Computer Society Conference on Computer Vision and
  Pattern Recognition}, volume~1, pages I--1090. Citeseer, 2001.

\bibitem{Batson2019}
Joshua Batson and Loic Royer.
\newblock {Noise2Self: Blind Denoising by Self-Supervision}.
\newblock In {\em The International Conference on Machine Learning (ICML)},
  2019.

\bibitem{briand2018}
Thibaud Briand, Gabriele Facciolo, and Javier Sánchez.
\newblock {Improvements of the Inverse Compositional Algorithm for Parametric
  Motion Estimation}.
\newblock {\em {Image Processing On Line}}, 8:435--464, 2018.

\bibitem{caelles2017osvos}
Sergi Caelles, Kevis-Kokitsi Maninis, Jordi Pont-Tuset, Laura Leal-Taix{\'e},
  Daniel Cremers, and Luc Van~Gool.
\newblock One-shot video object segmentation.
\newblock In {\em Proceedings of the IEEE conference on computer vision and
  pattern recognition}, pages 221--230, 2017.

\bibitem{18-chen-see-in-the-dark}
Chen Chen, Qifeng Chen, Jia Xu, and Vladlen Koltun.
\newblock Learning to see in the dark.
\newblock In {\em The IEEE Conference on Computer Vision and Pattern
  Recognition (CVPR)}, June 2018.

\bibitem{condat2012tvdemodeno}
Laurent Condat and Saleh Mosaddegh.
\newblock Joint demosaicking and denoising by total variation minimization.
\newblock In {\em 2012 19th IEEE International Conference on Image Processing},
  pages 2781--2784. IEEE, 2012.

\bibitem{Dabov2006}
Kostadin Dabov and Alessandro Foi.
\newblock {Image denoising with block-matching and 3D filtering}.
\newblock {\em Electronic Imaging}, 6064:1--12, 2006.

\bibitem{dabov2007bm3dcolor}
Kostadin Dabov, Alessandro Foi, Vladimir Katkovnik, and Karen~O Egiazarian.
\newblock Color image denoising via sparse 3d collaborative filtering with
  grouping constraint in luminance-chrominance space.
\newblock In {\em 2007 IEEE International Conference on Image Processing},
  volume~1, pages I -- 313--I -- 316, Sep. 2007.

\bibitem{danielyan2009bm3ddemodeno}
Aram Danielyan, Markku Vehvilainen, Alessandro Foi, Vladimir Katkovnik, and
  Karen Egiazarian.
\newblock Cross-color bm3d filtering of noisy raw data.
\newblock In {\em 2009 international workshop on local and non-local
  approximation in image processing}, pages 125--129. IEEE, 2009.

\bibitem{ehret2018model}
Thibaud Ehret, Axel Davy, Jean-Michel Morel, Gabriele Facciolo, and Pablo
  Arias.
\newblock Model-blind video denoising via frame-to-frame training.
\newblock In {\em Proceedings of the IEEE Conference on Computer Vision and
  Pattern Recognition (CVPR)}, June 2019.

\bibitem{Elad2006}
Michael Elad and Michal Aharon.
\newblock {Image denoising via sparse and redundant representations over
  learned dictionaries}.
\newblock {\em IEEE Transactions on Image Processing}, 15(12):3736--3745, 2006.

\bibitem{finn2017metalearning}
Chelsea Finn, Pieter Abbeel, and Sergey Levine.
\newblock Model-agnostic meta-learning for fast adaptation of deep networks.
\newblock In {\em Proceedings of the 34th International Conference on Machine
  Learning-Volume 70}, pages 1126--1135. JMLR. org, 2017.

\bibitem{getreuer2011color}
Pascal Getreuer.
\newblock Color demosaicing with contour stencils.
\newblock In {\em 2011 17th International Conference on Digital Signal
  Processing (DSP)}, pages 1--6. IEEE, 2011.

\bibitem{gharbi2016deep}
Micha{\"e}l Gharbi, Gaurav Chaurasia, Sylvain Paris, and Fr{\'e}do Durand.
\newblock Deep joint demosaicking and denoising.
\newblock {\em ACM Transactions on Graphics (TOG)}, 35(6):191, 2016.

\bibitem{godard2018deep}
Cl{\'e}ment Godard, Kevin Matzen, and Matt Uyttendaele.
\newblock Deep burst denoising.
\newblock In {\em Proceedings of the European Conference on Computer Vision
  (ECCV)}, pages 538--554, 2018.

\bibitem{gu2014weighted}
Shuhang Gu, Lei Zhang, Wangmeng Zuo, and Xiangchu Feng.
\newblock Weighted nuclear norm minimization with application to image
  denoising.
\newblock In {\em Proceedings of the IEEE Conference on Computer Vision and
  Pattern Recognition}, pages 2862--2869, 2014.

\bibitem{guo2018toward}
Shi Guo, Zifei Yan, Kai Zhang, Wangmeng Zuo, and Lei Zhang.
\newblock Toward convolutional blind denoising of real photographs.
\newblock {\em arXiv preprint arXiv:1807.04686}, 2018.

\bibitem{Hasinoff2016HDRplus}
Samuel~W. Hasinoff, Dillon Sharlet, Ryan Geiss, Andrew Adams, Jonathan~T.
  Barron, Florian Kainz, Jiawen Chen, and Marc Levoy.
\newblock {Burst photography for high dynamic range and low-light imaging on
  mobile cameras}.
\newblock {\em ACM Transactions on Graphics}, 35(6):1--12, nov 2016.

\bibitem{He2016}
Kaiming He, Xiangyu Zhang, Shaoqing Ren, and Jian Sun.
\newblock Deep residual learning for image recognition.
\newblock In {\em Proceedings of the IEEE conference on computer vision and
  pattern recognition (CVPR)}, pages 770--778, 2016.

\bibitem{Heide2014FlexISP}
Felix Heide, Karen Egiazarian, Jan Kautz, Kari Pulli, Markus Steinberger,
  Yun-Ta Tsai, Mushfiqur Rouf, Dawid Paj{\c{a}}k, Dikpal Reddy, Orazio Gallo,
  Jing Liu, and Wolfgang Heidrich.
\newblock {FlexISP}.
\newblock {\em ACM Transactions on Graphics}, 33(6):1--13, 11 2014.

\bibitem{hirakawa2006demodeno}
Keigo Hirakawa and Thomas~W Parks.
\newblock Joint demosaicing and denoising.
\newblock {\em IEEE Transactions on Image Processing}, 15(8):2146--2157, 2006.

\bibitem{Ioffe2015BatchShift}
Sergey Ioffe and Christian Szegedy.
\newblock Batch normalization: Accelerating deep network training by reducing
  internal covariate shift.
\newblock In Francis Bach and David Blei, editors, {\em Proceedings of the 32nd
  International Conference on Machine Learning}, volume~37 of {\em Proceedings
  of Machine Learning Research}, pages 448--456, Lille, France, 07--09 Jul
  2015. PMLR.

\bibitem{khashabi2014randomfields}
Daniel Khashabi, Sebastian Nowozin, Jeremy Jancsary, and Andrew~W Fitzgibbon.
\newblock Joint demosaicing and denoising via learned nonparametric random
  fields.
\newblock {\em IEEE Transactions on Image Processing}, 23(12):4968--4981, 2014.

\bibitem{klatzer2016demodeno}
Teresa Klatzer, Kerstin Hammernik, Patrick Knobelreiter, and Thomas Pock.
\newblock Learning joint demosaicing and denoising based on sequential energy
  minimization.
\newblock In {\em 2016 IEEE International Conference on Computational
  Photography (ICCP)}, pages 1--11. IEEE, 2016.

\bibitem{kokkinos2018cascade}
Filippos Kokkinos and Stamatios Lefkimmiatis.
\newblock Deep image demosaicking using a cascade of convolutional residual
  denoising networks.
\newblock In {\em The European Conference on Computer Vision (ECCV)}, September
  2018.

\bibitem{kokkinos2018resnet}
Filippos Kokkinos and Stamatios Lefkimmiatis.
\newblock Iterative residual network for deep joint image demosaicking and
  denoising.
\newblock {\em CoRR}, abs/1807.06403, 2018.

\bibitem{noise2void}
Alexander Krull, Tim{-}Oliver Buchholz, and Florian Jug.
\newblock Noise2void - learning denoising from single noisy images.
\newblock {\em CoRR}, abs/1811.10980, 2018.

\bibitem{kwan2019demosackingnasa}
Chiman Kwan, Bryan Chou, and James~F Bell~III.
\newblock Comparison of deep learning and conventional demosaicing algorithms
  for mastcam images.
\newblock {\em Electronics}, 8(3):308, 2019.

\bibitem{Lebrun2013}
Marc Lebrun, Antoni Buades, and Jean-Michel Morel.
\newblock {A Nonlocal Bayesian Image Denoising Algorithm}.
\newblock {\em SIAM Journal on Imaging Sciences}, 6(3):1665--1688, 2013.

\bibitem{lehtinen2018noise2noise}
Jaakko Lehtinen, Jacob Munkberg, Jon Hasselgren, Samuli Laine, Tero Karras,
  Miika Aittala, and Timo Aila.
\newblock Noise2noise: Learning image restoration without clean data.
\newblock {\em arXiv preprint arXiv:1803.04189}, 2018.

\bibitem{mildenhall2018burst}
Ben Mildenhall, Jonathan~T Barron, Jiawen Chen, Dillon Sharlet, Ren Ng, and
  Robert Carroll.
\newblock Burst denoising with kernel prediction networks.
\newblock In {\em Proceedings of the IEEE Conference on Computer Vision and
  Pattern Recognition}, pages 2502--2510, 2018.

\bibitem{morup2008csc}
Morten M{\o}rup, Mikkel~N Schmidt, and Lars~K Hansen.
\newblock Shift invariant sparse coding of image and music data.
\newblock {\em Submitted to Journal of Machine Learning Research}, 2008.

\bibitem{park2009denoisingBefore}
Sung~Hee Park, Hyung~Suk Kim, Steven Lansel, Manu Parmar, and Brian~A Wandell.
\newblock A case for denoising before demosaicking color filter array data.
\newblock In {\em 2009 Conference Record of the Forty-Third Asilomar Conference
  on Signals, Systems and Computers}, pages 860--864. IEEE, 2009.

\bibitem{plotz2017benchmarking}
Tobias Plotz and Stefan Roth.
\newblock Benchmarking denoising algorithms with real photographs.
\newblock In {\em Proceedings of the IEEE Conference on Computer Vision and
  Pattern Recognition}, pages 1586--1595, 2017.

\bibitem{Shocher2018zero-shot-SR}
Assaf Shocher, Nadav Cohen, and Michal Irani.
\newblock {Zero-Shot Super-Resolution Using Deep Internal Learning}.
\newblock In {\em The IEEE Conference on Computer Vision and Pattern
  Recognition (CVPR)}, 2018.

\bibitem{syu2018learning}
Nai-Sheng Syu, Yu-Sheng Chen, and Yung-Yu Chuang.
\newblock Learning deep convolutional networks for demosaicing.
\newblock {\em arXiv preprint arXiv:1802.03769}, 2018.

\bibitem{thevenaz1998pyramid}
Philippe Thevenaz, Urs~E Ruttimann, and Michael Unser.
\newblock A pyramid approach to subpixel registration based on intensity.
\newblock {\em IEEE transactions on image processing}, 7(1):27--41, 1998.

\bibitem{Tonioni2019Learning}
Alessio Tonioni, Oscar Rahnama, Thomas Joy, Luigi {Di Stefano}, Thalaiyasingam
  Ajanthan, and Philip Torr.
\newblock {Learning to Adapt for Stereo}.
\newblock In {\em The IEEE Conference on Computer Vision and Pattern
  Recognition (CVPR)}, 2019.

\bibitem{Tonioni2018RealtimeStereo}
Alessio Tonioni, Fabio Tosi, Matteo Poggi, Stefano Mattoccia, and Luigi {Di
  Stefano}.
\newblock {Real-time self-adaptive deep stereo}.
\newblock In {\em The IEEE Conference on Computer Vision and Pattern
  Recognition (CVPR)}, 2019.

\bibitem{ulyanov2018deep}
Dmitry Ulyanov, Andrea Vedaldi, and Victor Lempitsky.
\newblock Deep image prior.
\newblock In {\em Proceedings of the IEEE Conference on Computer Vision and
  Pattern Recognition}, pages 9446--9454, 2018.

\bibitem{Wronski2019}
Bartlomiej Wronski, Ignacio Garcia-Dorado, Manfred Ernst, Damien Kelly, Michael
  Krainin, Chia-Kai Liang, Marc Levoy, and Peyman Milanfar.
\newblock {Handheld multi-frame super-resolution}.
\newblock {\em ACM Transactions on Graphics}, 38(4):1--18, jul 2019.

\bibitem{yu2012ple}
Guoshen Yu, G. Sapiro, and S. Mallat.
\newblock Solving inverse problems with piecewise linear estimators: From
  gaussian mixture models to structured sparsity.
\newblock {\em Image Processing, IEEE Transactions on}, 21(5):2481--2499, May
  2012.

\bibitem{Zhang2017BeyondDenoising}
Kai Zhang, Wangmeng Zuo, Yunjin Chen, Deyu Meng, and Lei Zhang.
\newblock {Beyond a Gaussian Denoiser: Residual Learning of Deep CNN for Image
  Denoising}.
\newblock {\em IEEE Transactions on Image Processing}, 26(7):3142--3155, 7
  2017.

\bibitem{17-zhang-ffdnet}
Kai Zhang, Wangmeng Zuo, and Lei Zhang.
\newblock {FFDNet: Toward a Fast and Flexible Solution for {\{}CNN{\}} based
  Image Denoising}.
\newblock {\em CoRR}, abs/1710.0, 2017.

\bibitem{Zhao2017LossNetworks}
Hang Zhao, Orazio Gallo, Iuri Frosio, and Jan Kautz.
\newblock {Loss Functions for Image Restoration With Neural Networks}.
\newblock {\em IEEE Transactions on Computational Imaging}, 3(1):47--57, 3
  2017.

\bibitem{zhussip2019n2nclipping}
Magauiya Zhussip, Shakarim Soltanayev, and Se~Young Chun.
\newblock Theoretical analysis on noise2noise using stein's unbiased risk
  estimator for gaussian denoising: Towards unsupervised training with clipped
  noisy images.
\newblock {\em CoRR}, abs/1902.02452, 2019.

\end{thebibliography}
}

\end{document}